\documentclass[fleqn,10pt]{wlscirep}
\usepackage[utf8]{inputenc}
\usepackage[T1]{fontenc}
\usepackage{amsmath,amssymb,amsfonts}
\usepackage{longtable}
\usepackage{hyperref}
\usepackage{xurl}
\usepackage{tabularray}
\title{Comparative Analysis of 47 Context-Based Question Answer Models Across 8 Diverse Datasets}

\author[1,2]{Muhammad Muneeb}
\author[1,2,*]{David B. Ascher}
\author[3]{Ahsan Baidar Bakht}
\affil[1]{School of Chemistry and Molecular Biology, The University of Queensland, Brisbane, 4067, Australia}
\affil[2]{Computational Biology and Clinical Informatics, Baker Heart and Diabetes Institute, Melbourne, 3004, Australia}
\affil[3]{Mechanical Engineering Department, Khalifa University, Abu Dhabi, UAE}

\affil[*]{d.ascher@uq.edu.au}


\begin{abstract}
 Context-based question answering (CBQA) models provide more accurate and relevant answers by considering the contextual information. They effectively extract specific information given a context, making them functional in various applications involving user support, information retrieval, and educational platforms. In this manuscript, we benchmarked the performance of 47 CBQA models from Hugging Face on eight different datasets. This study aims to identify the best-performing model across diverse datasets without additional fine-tuning. It is valuable for practical applications where the need to retrain models for specific datasets is minimized, streamlining the implementation of these models in various contexts. The best-performing models were trained on the SQuAD v2 or SQuAD v1 datasets. The best-performing model was ahotrod/electra\_large\_discriminator\_squad2\_512, which yielded 43\% accuracy across all datasets. We observed that the computation time of all models depends on the context length and the model size. The model's performance usually decreases with an increase in the answer length. Moreover, the model's performance depends on the context complexity. We also used the Genetic algorithm to improve the overall accuracy by integrating responses from other models. ahotrod/electra\_large\_discriminator\_squad2\_512 generated the best results for bioasq10b-factoid (65.92\%), biomedical\_cpgQA (96.45\%), QuAC (11.13\%), and Question Answer Dataset (41.6\%). Bert-large-uncased-whole-word-masking-finetuned-squad achieved an accuracy of 82\% on the IELTS dataset. Palak/microsoft\_deberta-large\_squad accuracy was 31\% on the JournalQA dataset. Twmkn9/albert-base-v2-squad2 was the best-performing model for the ScienceQA dataset with an accuracy of 24.6\%. This study contributes to the broader goal of optimizing the use of CBQA models across diverse datasets by providing insights into the impact of context complexity, answer length, and question types on model performance.
\end{abstract}
\begin{document}

\flushbottom
\maketitle
%
%
\thispagestyle{empty}


\section*{Introduction}
In the era of information abundance, the need for efficient extraction and retrieval of relevant information from vast textual repositories has become paramount \cite{Landhuis2016,Alanazi2021}. Question-answering (QA) models are pivotal in facilitating human-machine interaction by interpreting complex linguistic structures, discerning contextual nuances, and generating accurate responses \cite{sen2020models}. QA tasks are categorized into three classes: Answer generation, where the answer is generated based on the question; answer selection, where models choose answers from multiple options; and answer extraction, where models extract answers from the context \cite{allam2012question}. We considered models that use a question and context to extract an answer.

A context-based question answering (CBQA) model is a type of natural language processing (NLP) model designed to understand and respond to questions in the context of a given passage or document \cite{Liddy2004,dsas,Li2020,Li2017,fssss}. The exigency for QA models is underscored by their applications across diverse domains, including information retrieval \cite{re}, customer support, education \cite{finnie2022robots}, and medical literature \cite{Lee2019,Cao2011,Cao2010}, QA models also aid in extracting information from research articles, enhancing the efficiency of literature reviews and evidence-based decision-making \cite{Brown}. 

CBQA models initiate their process with tokenization \cite{song2020fast} that combines the question with the context, and the data is then segmented into smaller units known as tokens. These tokens are represented as vectors in a high-dimensional space through embedding, encompassing token, position, and segment embedding (capturing the semantic meaning and relationships between words) \cite{bai2021segatron}. The embeddings are passed to the transformer, which uses a self-attention mechanism \cite{attention} to capture dependencies between words, irrespective of their position in the sequence. Subsequently, QA models undergo pre-training on extensive datasets with a specific objective \cite{wang2022language}. For instance, in the Masked Language Model (MLM) objective \cite{salazar2019masked}, a portion of tokens is masked (replaced with a unique token called MASK), and the model predicts the correct token in place of MASK to acquire contextualized language representations \cite{bert}. This pretraining phase aids the model in developing a broad understanding of language. Following pretraining \cite{pretrain}, the model undergoes fine-tuning \cite{dodge2020fine} on a task-specific dataset for QA. The question-answering head is added at this stage for fine-tuning the QA datasets, which predicts the answer's start and end token.  

The evaluation of CBQA models typically involves metrics such as Exact Match (EM) and F1 score, assessing the model's ability to generate the exact answer span or a close approximation \cite{risch2021semantic,chen2019evaluating}. Additionally, researchers are exploring more nuanced evaluation methods that consider aspects such as answer relevance, coherence, and informativeness \cite{Rodrigo2017}.

Benchmarking question-answering (QA) models is critical to assess and compare their performance systematically, providing a basis for understanding their capabilities and limitations across diverse tasks and datasets \cite{wang2022modern,Usbeck2019}. Benchmarking also serves to identify areas of improvement and challenges in QA models. The diverse nature of benchmarks allows for a comprehensive evaluation of models' performance in different domains, contextual complexities, and linguistic nuances. Such evaluations aid researchers in identifying strengths and weaknesses, refining existing models, and inspiring the development of novel methodologies.

Hugging Face is a popular platform for natural language processing models \cite{huggingface}. It hosts an extensive repository of question-answering models, including transformer-based models like BERT \cite{bert}, GPT, and RoBERTa \cite{roberta}, Dilbert \cite{DistilBERT}, ELECTRA \cite{ELECTRA} and T5 \cite{T5}. Pre-trained on large corpora, these models have demonstrated state-of-the-art performance across a spectrum of NLP tasks, including question-answering. We considered 47 context-based question-answer models from the Hugging Face.

This study explored several crucial aspects of model performance and efficiency across diverse datasets. Our critical investigations encompassed (1) Identifying the top-performing models based on accuracy and computation time across eight distinct datasets, (2) Analyzing the impact of model size and context length on execution time, (3) Investigating the relationship between context complexity and model performance, (4) Assessing the impact of answer length on the accuracy of the model, providing insights into how different answer lengths influence the overall effectiveness of the model, (5) Examining the influence of answer types on model performance, offering valuable perspectives on how diversity of answers challenge the model's predictive capabilities, and (6) Implementing an ensemble-based approach that leverages genetic algorithms to discover optimal combinations of models, resulting in superior overall performance.

We comprehensively explored critical factors influencing model behavior, offering a nuanced understanding of their collective impact on computation time and overall performance.

\section*{Results}
\begin{figure}[!ht]
\resizebox{1.1\textwidth}{!}{%
     \includegraphics[width=15cm,height=12cm]{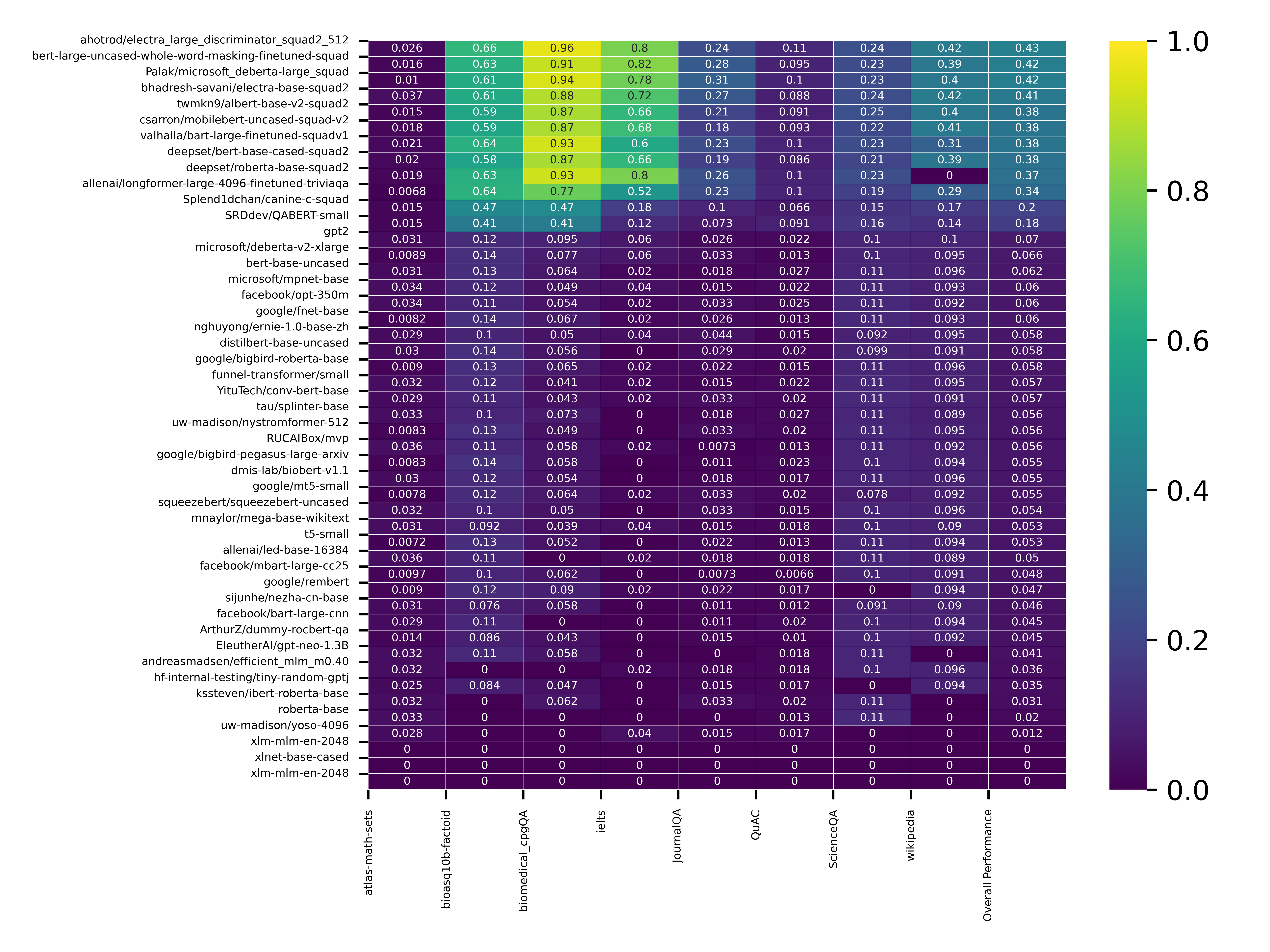}}
     \caption{\textbf{This heatmap illustrates the performance of each model across eight datasets.} We computed the overall performance by summing the model performance for a specific dataset and dividing it by the number of datasets. We sorted the entries in the heatmap based on overall performance.}
\label{heatmap1}
\end{figure}

Among the top ten models, overall performance exceeded 30 percent \ref{heatmap1}. While many models exhibited similar overall performance, models' performance on individual datasets varied. Top-performing models were trained on the SQuAD v2 or SQuAD v1 datasets \ref{topperformer}.

\begin{table}[!ht]
\centering
\resizebox{1.1\textwidth}{!}{%
\begin{tabular}{|l|l|l|l|l|}
\hline
 & \textbf{Model Name} & \textbf{Model Size (MB)} & \textbf{Dataset} & \textbf{Overall Performance} \\ \hline
1 & ahotrod/electra\_large\_discriminator\_squad2\_512 & 319 & SQuAD v2 & 0.432683 \\ \hline
2 & bert-large-uncased-whole-word-masking-finetuned-squad  \cite{DBLP:journals/corr/abs-1810-04805} & 320 & BookCorpus and English Wikipedia & 0.422568 \\ \hline
3 & Palak/microsoft\_deberta-large\_squad & 387 & SQuAD v1 & 0.422074 \\ \hline
4 & bhadresh-savani/electra-base-squad2 & 104 & SQuAD v2 & 0.407044 \\ \hline
5 & twmkn9/albert-base-v2-squad2 & 12 & SQuAD v2 & 0.384236 \\ \hline
6 & csarron/mobilebert-uncased-squad-v2 & 24 & SQuAD v2 & 0.383692 \\ \hline
7 & valhalla/bart-large-finetuned-squadv1 & 388 & SQuAD v1 & 0.383498 \\ \hline
8 & deepset/bert-base-cased-squad2 & 104 & SQuAD v2 & 0.37514 \\ \hline
9 & deepset/roberta-base-squad2 & 119 & SQuAD v2 & 0.370862 \\ \hline
\end{tabular}%
}
\caption{This table shows information about the top 9 models, including the model name, size in megabytes, the dataset used for training, and the overall performance.}
\label{topperformer}
\end{table}

The second-best performer, Bert-large-uncased-whole-word-masking-finetuned-squad, was trained on BookCorpus and English Wikipedia. Other algorithms trained on the same dataset did not yield better results, potentially due to differences in model sizes. For instance, this model had a size of 320 MB, while other models trained on the same dataset had a size below 100 MB, resulting in low accuracy.

\begin{figure}[!ht]
\resizebox{1.1\textwidth}{!}{%
     \includegraphics[width=15cm,height=14cm]{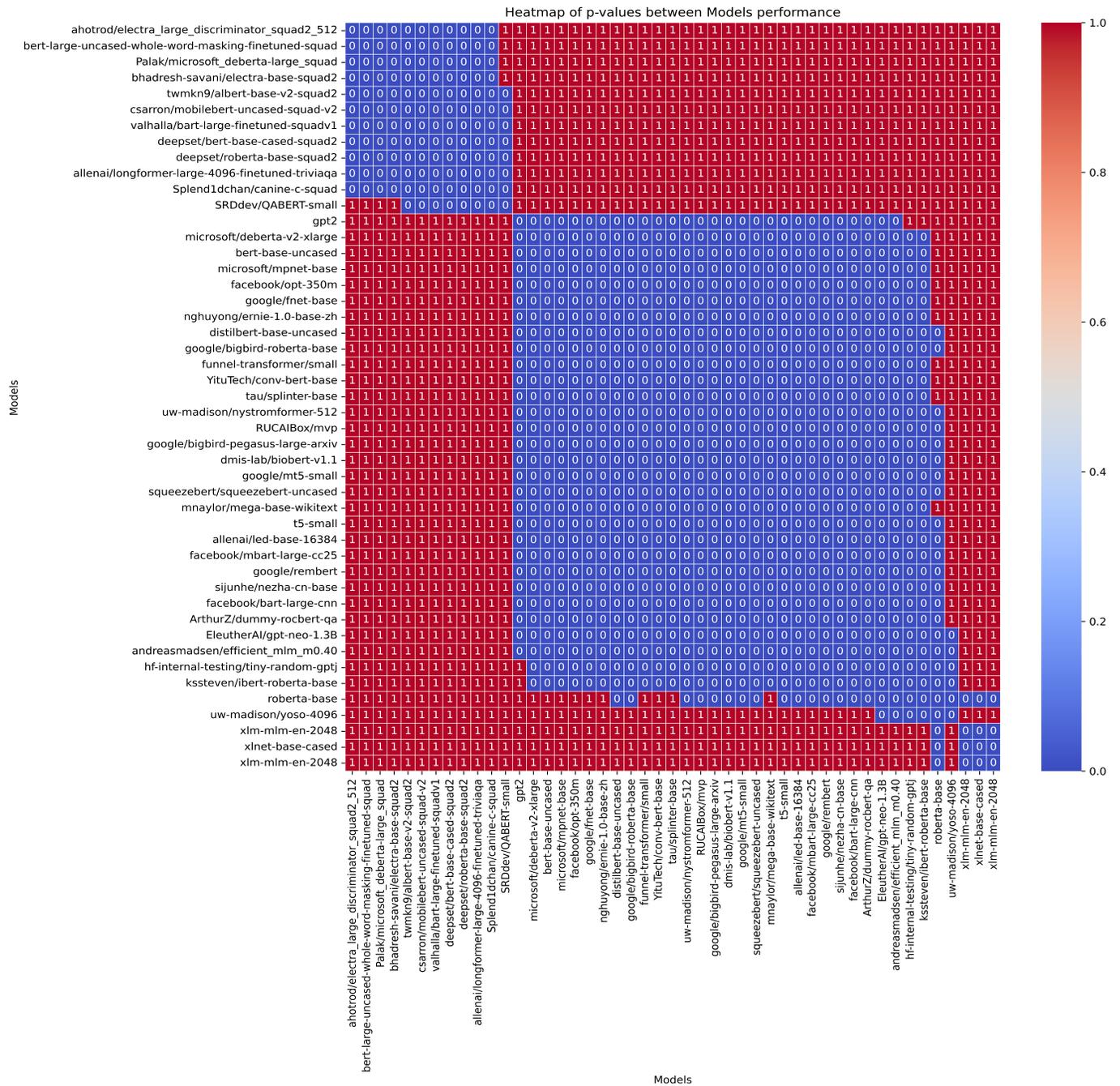}}
     \caption{\textbf{This heatmap depicts the t-test comparing the performance of each model across datasets.} We sorted the values based on performance and calculated the t-test (p-value < 0.1). If the p-value for a t-test was less than 0.1, we assigned a value of 1 to that particular entry; otherwise, it is marked as 0. This representation indicates whether a specific pair of two models is statistically significant.}
\label{pvalue1}
\end{figure}

Table \ref{topperformer} highlights the top-performing models. ahotrod/electra\_large\_discriminator\_squad2\_512 achieved an overall accuracy of 43\%, with a size of 319 MB. In contrast, twmkn9/albert-base-v2-squad2 achieved an accuracy of 38\% with a significantly smaller size of 12 MB. Model size indeed impacted performance; however, the base model used for training also played a crucial role. Models trained on various base models (Electra, Albert, MobileBERT, BERT, and Roberta) exhibited performance in descending order, as shown in Table \ref{topperformer}. We conducted a t-test (p-value < 0.1) to evaluate the difference in the performance of the models. It involved comparing the performance of each model against every other model across eight datasets. The results indicate no statistically significant difference among the top 11 best-performing models \ref{pvalue1}.

\begin{figure}[!ht]
\resizebox{1.1\textwidth}{!}{%
     \includegraphics[width=15cm,height=12cm]{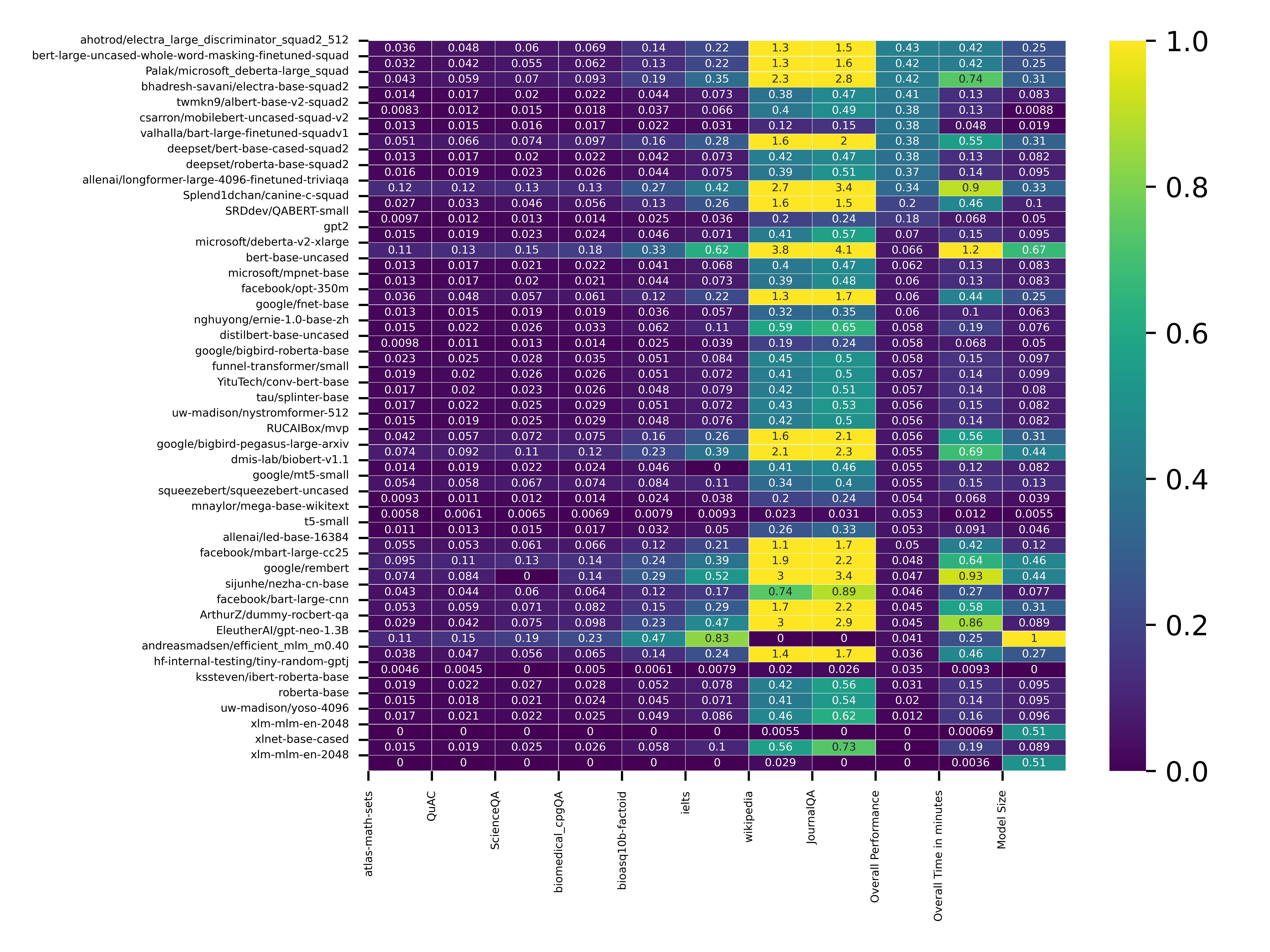}}
     \caption{\textbf{This heat map illustrates the average execution time in minutes for each model across all datasets, overall execution time, overall performance, and the respective model sizes.} The average execution time for a specific model on a given dataset was calculated by dividing the model execution time on a specific dataset by the number of questions in that dataset. This calculated average provides valuable insight into visualizing the impact of context length on execution time. Model sizes were normalized between 0 and 1 using min-max normalization.}
\label{step2}
\end{figure}

We observed that the context length and the model size have a noticeable impact on the model execution time. The correlation between the overall execution time of the models and their size across all datasets was 51\%, indicating a significant increase in execution time with larger models. However, the correlation between the overall execution time and overall performance across all datasets was 13\%, suggesting that longer execution time or larger models may not necessarily improve the performance. It aligns with previous inferences, emphasizing the importance of selecting datasets and the base model for training to achieve better results. There was a correlation of -8\% between overall performance and model size. The negative correlation implies that an increase in model size does not necessarily lead to an increase in overall performance.
Examining specific models, the first (ahotrod/electra\_large\_discriminator\_squad2\_512) and fourth (bhadresh-savani/electra-base-squad2) models \ref{step2} demonstrated comparable performance (approximately 0.43 and 0.41, respectively). However, a substantial difference in execution time was observed due to the varying sizes of the models, which were 319 MB and 104 MB, respectively. In scenarios where rapid results are required and a compromise on performance is acceptable, opting for the smaller-sized model is advisable.

\begin{figure}[!ht]
\resizebox{1.1\textwidth}{!}{%
     \includegraphics[width=15cm,height=14cm]{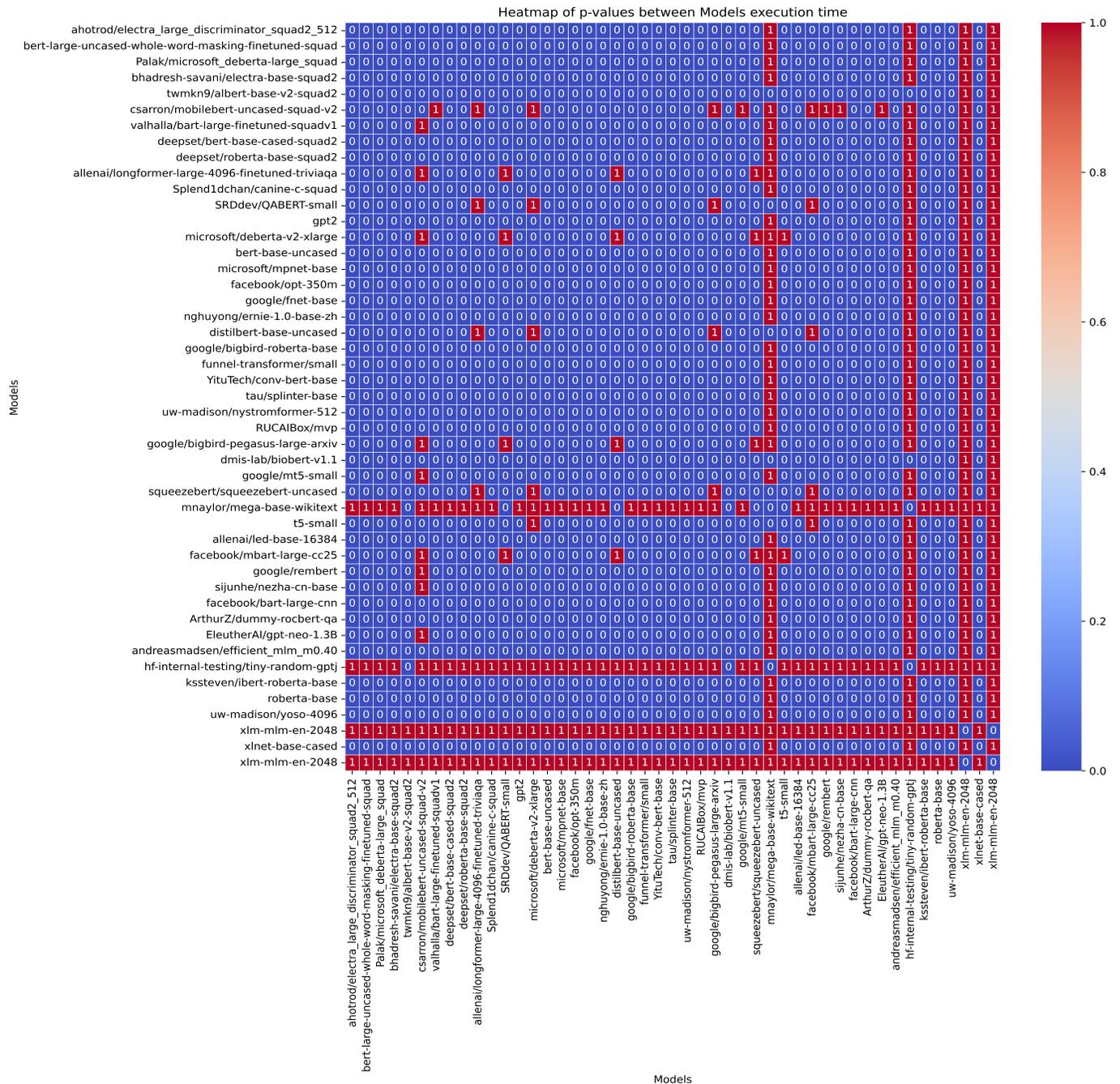}}
     \caption{\textbf{This heatmap depicts the t-test comparing the execution time of each model across datasets.} We sorted the values based on performance and calculated the t-test (p-value < 0.1). If the p-value for a t-test was less than 0.1, we assigned a value of 1 to that particular entry; otherwise, it is marked as 0. This representation indicates whether a specific pair of two models is statistically significant.}
\label{pvalue2}
\end{figure}

We conducted a t-test (p-value < 0.1) to evaluate the difference in the computation time of the models. It involved comparing the computation time of each model against every other model across eight datasets. The results indicate no statistically significant difference among the top 6 best-performing models \ref{pvalue2}.

We calculated the correlation between the model execution time on eight datasets and overall performance (Figure: \ref{step2} and Table \ref{executioninference}). There was an increase in correlation between performance and model execution time, suggesting more execution time can bring better performance; however, this correlation might be coincidental. To further explore, we computed the correlation between the model execution time for each dataset and the model size. Initially, there was an increase, followed by a decrease as the context length increased. It implies that the model size alone does not explain the execution time; the context length also plays a significant role. If this correlation had increased linearly, it would have suggested that context length does not impact model performance and that only the model size influences execution time. Additionally, we examined the correlation between the model execution time for each dataset and overall execution time, revealing that as the context length increases, the execution time also increases. All these observations are shown in Table \ref{executioninference}.

\begin{table}[!ht]
\resizebox{1.1\textwidth}{!}{%
\begin{tabular}{|l|l|l|l|l|l|}
\hline
 & \textbf{Model Name} & \textbf{Average Context Length} & \textbf{Overall Performance} & \textbf{Model Size (MB)} & \textbf{Overall Time in Minutes} \\ \hline
1 & atlas-math-sets & 3 & 0.04 & 0.72 & 0.78 \\ \hline
2 & QuAC & 84 & 0.05 & 0.77 & 0.78 \\ \hline
3 & ScienceQA & 147 & 0.07 & 0.71 & 0.68 \\ \hline
4 & biomedical\_cpgQA & 161 & 0.04 & 0.78 & 0.8 \\ \hline
5 & bioasq10b-factoid & 351 & 0.05 & 0.78 & 0.81 \\ \hline
6 & IELTS & 907 & 0.04 & 0.77 & 0.82 \\ \hline
7 & Question Answer Dataset & 5227 & 0.13 & 0.42 & 0.99 \\ \hline
8 & JournalQA & 5568 & 0.16 & 0.43 & 0.99 \\ \hline
\end{tabular}%
}
\caption{This table depicts the correlation between the execution time of all models for a specific dataset and the overall performance, model size, and total execution time for each dataset. The average context length is presented in the second column.}
\label{executioninference}
\end{table}

\begin{figure}[!ht]
\resizebox{1.1\textwidth}{!}{%
     \includegraphics[width=15cm,height=12cm]{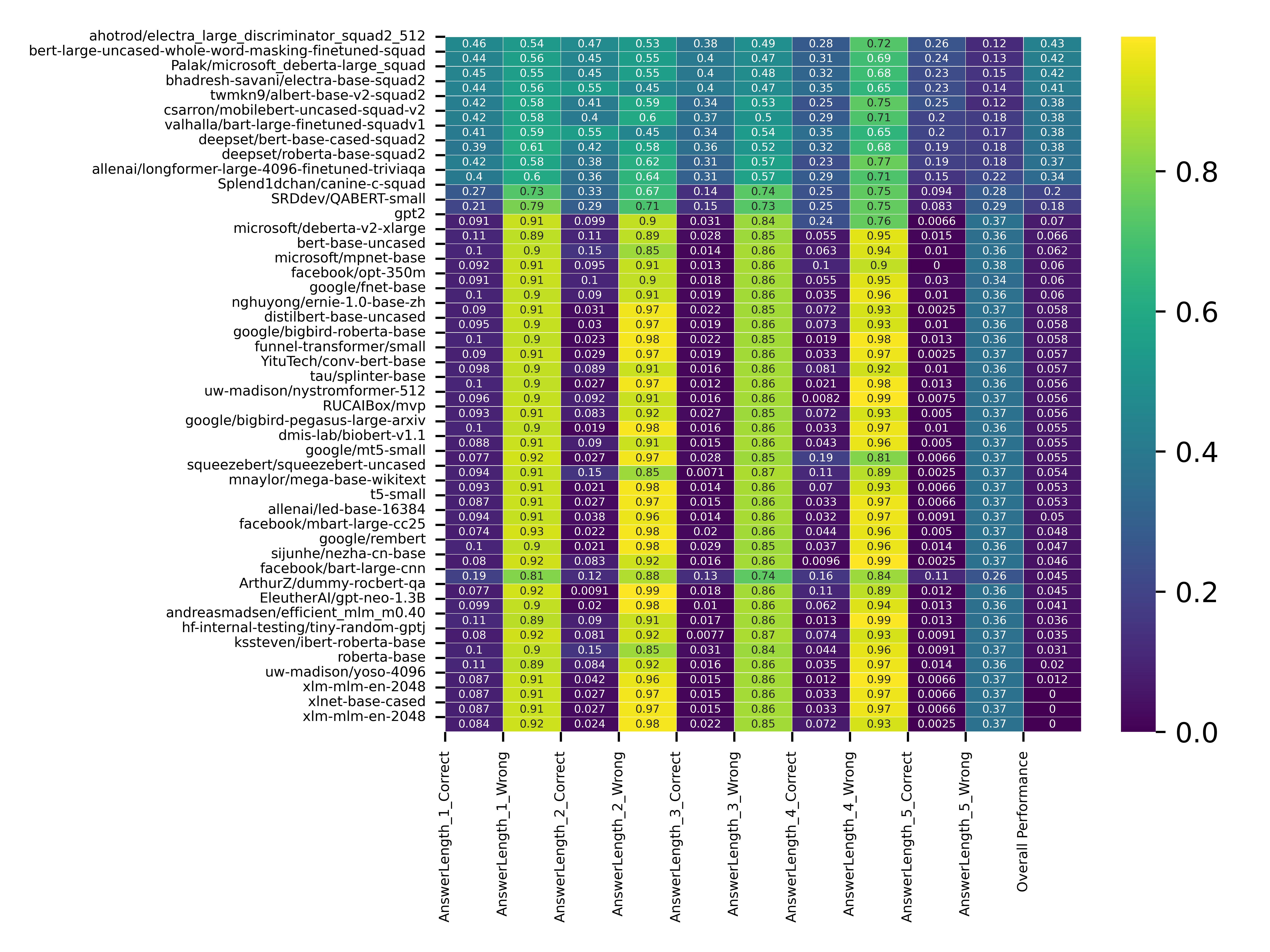}}
    \caption{\textbf{This heatmap illustrates the impact of answer length on model performance.} As the question length increases, the models' performance may either decrease or increase. The answer length spans from 1 to 5 words, and we calculated the percentage of correctly and incorrectly classified answers by each model for each answer length across all datasets. This information is then utilized to generate the heatmap.}
\label{step7}
\end{figure}

We analyzed the impact of answer length on model performance \ref{step7}. Our findings revealed that the performance of the top-performing model could exhibit variations with changes in answer length. Specifically, we identified a consistent decline in model performance as the answer length increased. For instance, the model ahotrod/electra\_large\_discriminator\_squad2\_512 emerged as the best performer overall. However, when predicting answers consisting of only two words, the model bhadresh-savani/electra-base-squad2 demonstrated a noteworthy accuracy of 55\%, outperforming the best model, which achieved an accuracy of 0.47\%.

Table \ref{answereffect} presents the models that yielded the best results across various answer lengths for all datasets. The values are displayed in pairs, following the format [Answer length, Best Accuracy]. The optimal performing model can vary based on the length of the answer. When predicting an answer of a specific length, it becomes imperative to identify which model is likely to produce the best results for the answer of a specific length. This variation is influenced by factors such as model architecture and other design considerations. Some models are tailored to pinpoint the precise answer, while others are designed to return more contextual information around the actual answer, resulting in a more detailed response.

\begin{table}[!ht]
\hspace{-1cm} 
\resizebox{1.1\textwidth}{!}{%
\begin{tabular}{|l|l|l|l|l|l|l|l|l|l|}
\hline
 & \textbf{models} & \textbf{atlas-math-sets} & \textbf{bioasq10b-factoid} & \textbf{biomedical\_cpgQA} & \textbf{ielts} & \textbf{JournalQA} & \textbf{QuAC} & \textbf{ScienceQA} & \textbf{Question Answer Dataset} \\ \hline
\textbf{1} & \textbf{twmkn9/albert-base-v2-squad2} & - & - & - & - & - & - & {[}{[}1, 0.36{]}, {[}3, 0.08{]}{]} & - \\ \hline
\textbf{2} & \textbf{valhalla/bart-large-finetuned-squadv1} & {[}{[}4, 1.0{]}{]} & {[}{[}1, 0.74{]}{]} & {[}{[}5, 1.0{]}{]} & - & - & - & {[}{[}2, 0.21{]}{]} & - \\ \hline
\textbf{3} & \textbf{deepset/roberta-base-squad2} & - & {[}{[}3, 0.58{]}{]} & - & {[}{[}2, 0.94{]}{]} & - & - & - & - \\ \hline
\textbf{4} & \textbf{Palak/microsoft\_deberta-large\_squad} & - & - & {[}{[}3, 1.0{]}{]} & - & {[}{[}3, 0.18{]}, {[}4, 0.43{]}{]} & - & - & - \\ \hline
\textbf{5} & \textbf{bhadresh-savani/electra-base-squad2} & {[}{[}1, 0.04{]}, {[}2, 1.0{]}{]} & - & - & - & {[}{[}2, 0.43{]}{]} & - & - & {[}{[}1, 0.31{]}, {[}5, 0.57{]}{]} \\ \hline
\textbf{6} & \textbf{gpt2} & - & - & - & {[}{[}4, 1.0{]}{]} & - & - & - & - \\ \hline
\textbf{7} & \textbf{allenai/longformer-large-4096-finetuned-triviaqa} & - & {[}{[}2, 0.65{]}{]} & - & - & - & {[}{[}3, 0.1{]}, {[}4, 0.05{]}{]} & - & - \\ \hline
\textbf{8} & \textbf{csarron/mobilebert-uncased-squad-v2} & - & - & {[}{[}1, 0.98{]}{]} & - & - & - & - & {[}{[}3, 0.63{]}{]} \\ \hline
\textbf{9} & \textbf{SRDdev/QABERT-small} & - & - & - & - & - & - & {[}{[}4, 0.18{]}{]} & - \\ \hline
\textbf{10} & \textbf{bert-large-uncased-whole-word-masking-finetuned-squad} & - & - & - & {[}{[}3, 0.92{]}{]} & {[}{[}1, 0.27{]}{]} & - & - & - \\ \hline
\textbf{11} & \textbf{ahotrod/electra\_large\_discriminator\_squad2\_512} & {[}{[}3, 0{]}, {[}5, 0{]}{]} & {[}{[}4, 0.47{]}, {[}5, 0{]}{]} & {[}{[}2, 0.95{]}, {[}4, 0.95{]}{]} & {[}{[}1, 0.89{]}, {[}5, 0{]}{]} & {[}{[}5, 0.6{]}{]} & {[}{[}1, 0.23{]}, {[}2, 0.09{]}, {[}5, 0{]}{]} & {[}{[}5, 0{]}{]} & {[}{[}2, 0.74{]}, {[}4, 0.61{]}{]} \\ \hline
\end{tabular}%
}
\caption{This table illustrates the comparative performance of different models for each answer length. The values are presented in the format [Answer Length, Best Accuracy]. "-" denotes that a particular model did not emerge as the best performer for any specific answer length.}
\label{answereffect}
\end{table}

\begin{figure}[!ht]
\hspace{-1cm} 
\resizebox{1.1\textwidth}{!}{%
    \includegraphics[width=20cm,height=17cm]{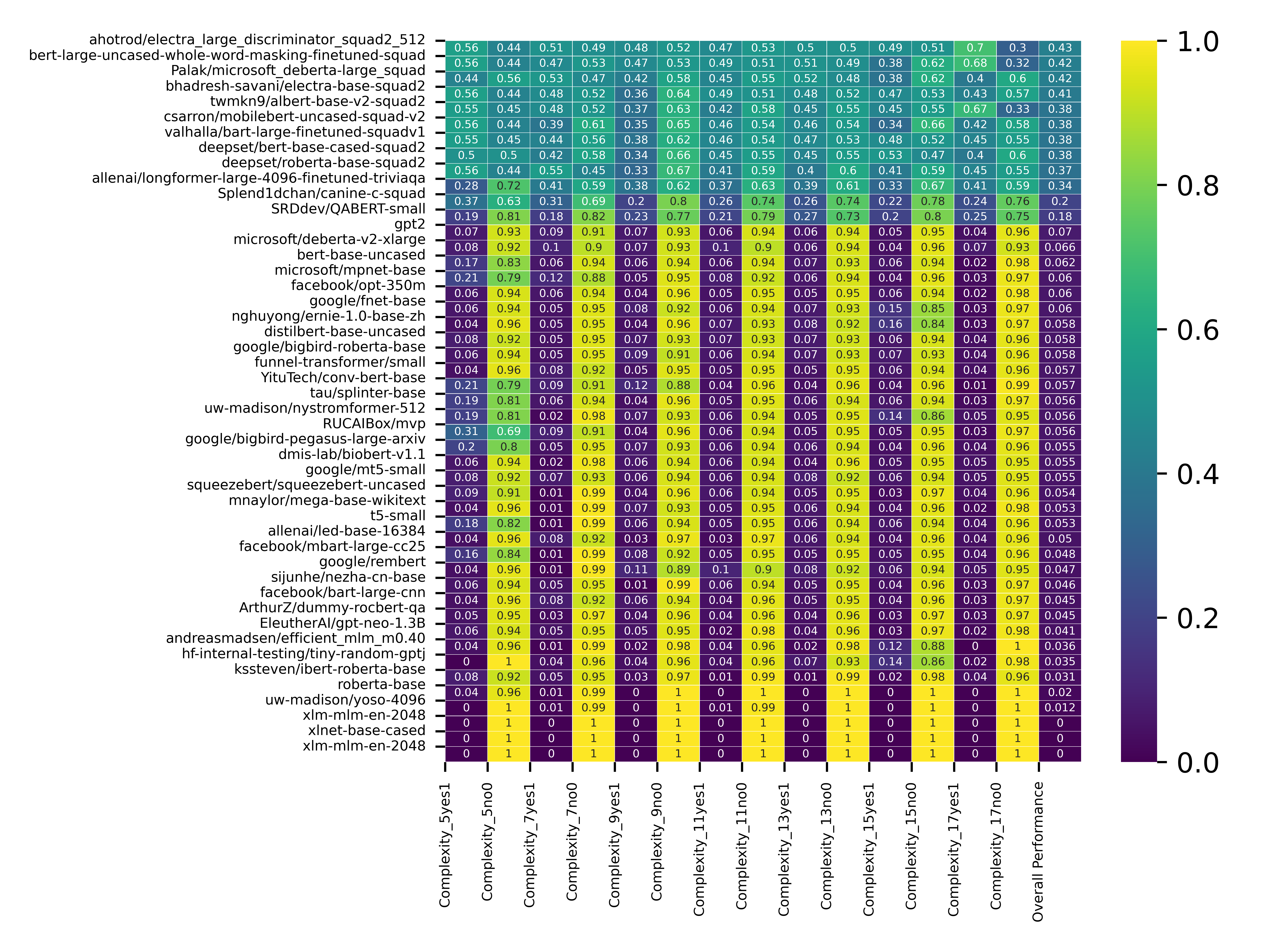}}
    \caption{\textbf{This heatmap illustrates the impact of complexity on model performance.} As the complexity increases, the models' performance may either decrease or increase. The complexity measure spans from 5 to 17, with an interval of 2. No specific pattern was observed for this analysis.}
\label{step9}
\end{figure}

We conducted a thorough analysis to investigate context complexity's influence on the performance of models. Our findings reveal a notable correlation between varying context complexities and the deviations in model performance \ref{step9}. However, intriguingly, we did not find any discernible pattern regarding the specific impact of context complexity on model performance. Further exploration is needed to unravel the nuanced dynamics that contribute to the observed variations in performance.

We employed a genetic algorithm to discover combinations of models that could enhance overall performance. We aimed to generate a more robust and powerful model by leveraging the weaker models. While a combination of models can obtain optimal performance, the improvement is not statistically significant. In many instances, the performance gains achieved by combining models were either on par with the best individual model or marginally less, at the expense of increased computation time.

\begin{figure}[!ht]
\hspace{-1cm} 
\resizebox{1.1\textwidth}{!}{%
     \includegraphics[width=20cm,height=17cm]{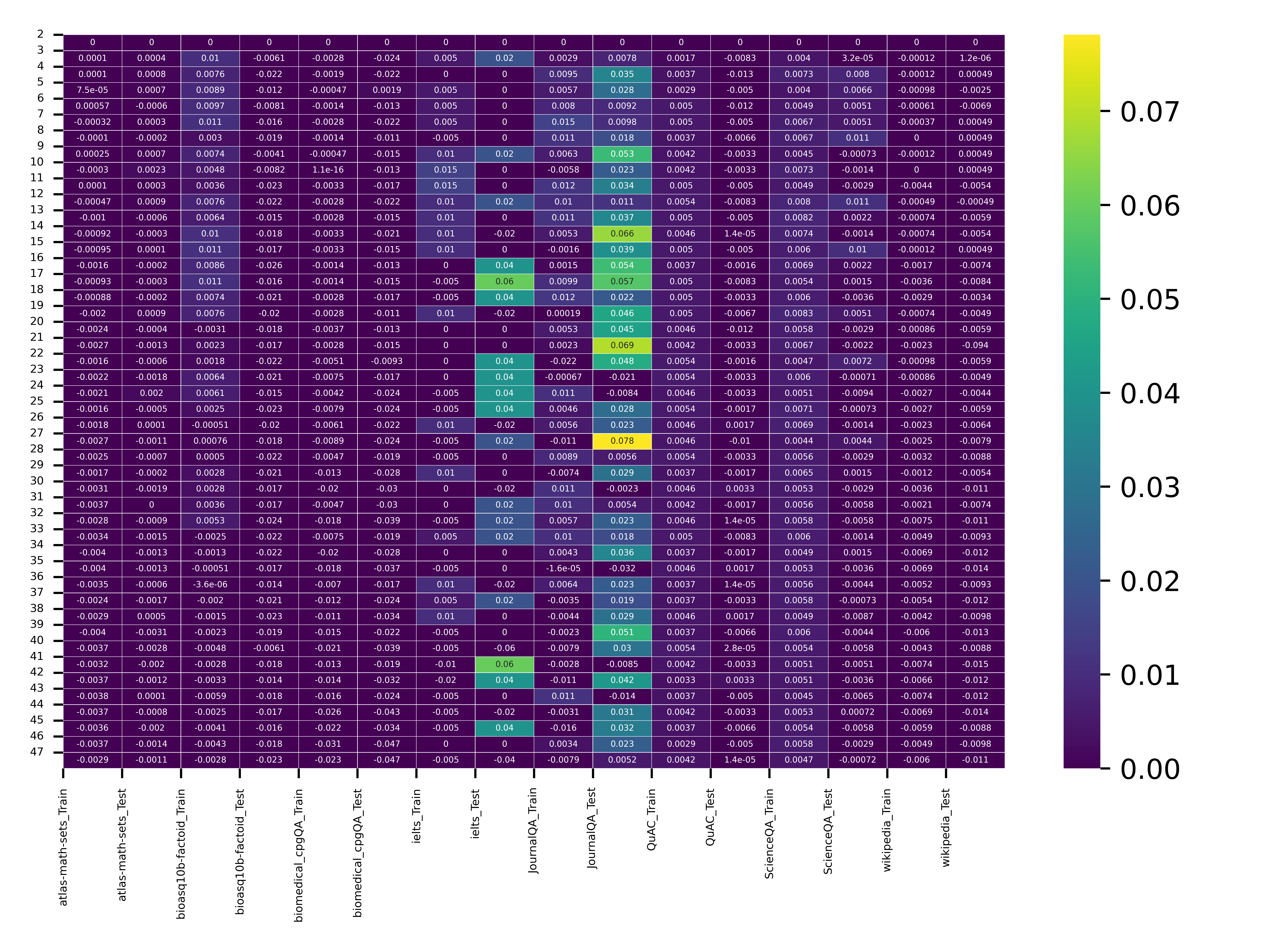}}
     \caption{\textbf{The heatmap represents the performance improvements achieved by the genetic algorithm when optimizing performance using a combination of models.}  Each cell displays the averaged five-fold training and test accuracy for each dataset across combinations of 2 to 47 models shown on the y-axis. The heatmap illustrates that, on the whole, there was no substantial improvement in performance.}
\label{step10}
\end{figure}

To conduct our analysis, we employed a 5-fold cross-validation approach to generate training and test sets. The genetic algorithm was then utilized to optimize the number of models included in the optimal configuration, ranging from 2 to 47 models. The genetic algorithm selected the model with the highest answer score, optimizing the accuracy of the training data. Subsequently, we evaluated the performance of the combined model on the test set and reported improvements compared to the baseline \ref{step10}.

The JournalQA dataset comprised questions derived from two papers, one centered on solar power prediction and the other on the polygenic risk associated with diabetes. Our observations revealed that the effectiveness of the question-answer model varied depending on the type of information extracted from the papers. To simplify this diversity, we categorized information into ten groups: Type, Performance, Location, Time, Date, Features, Output, Evaluation metric, Best Model, Number, and Dataset. The performance category means the optimal achievements of specific algorithms, while location details the specific geographical information. Time means date ranges. Output means the evaluation metrics or output of the system discussed in the paper, and evaluation metrics detail the specific metrics. The best model identified the highest-performing model, the number conveyed numerical information, and the data specified the dataset used in the specific study. Lastly, the type category encompassed anything related to categorization.

\begin{figure}[!ht]
\hspace{-1cm} 
\resizebox{1.1\textwidth}{!}{%
     \includegraphics[width=20cm,height=17cm]{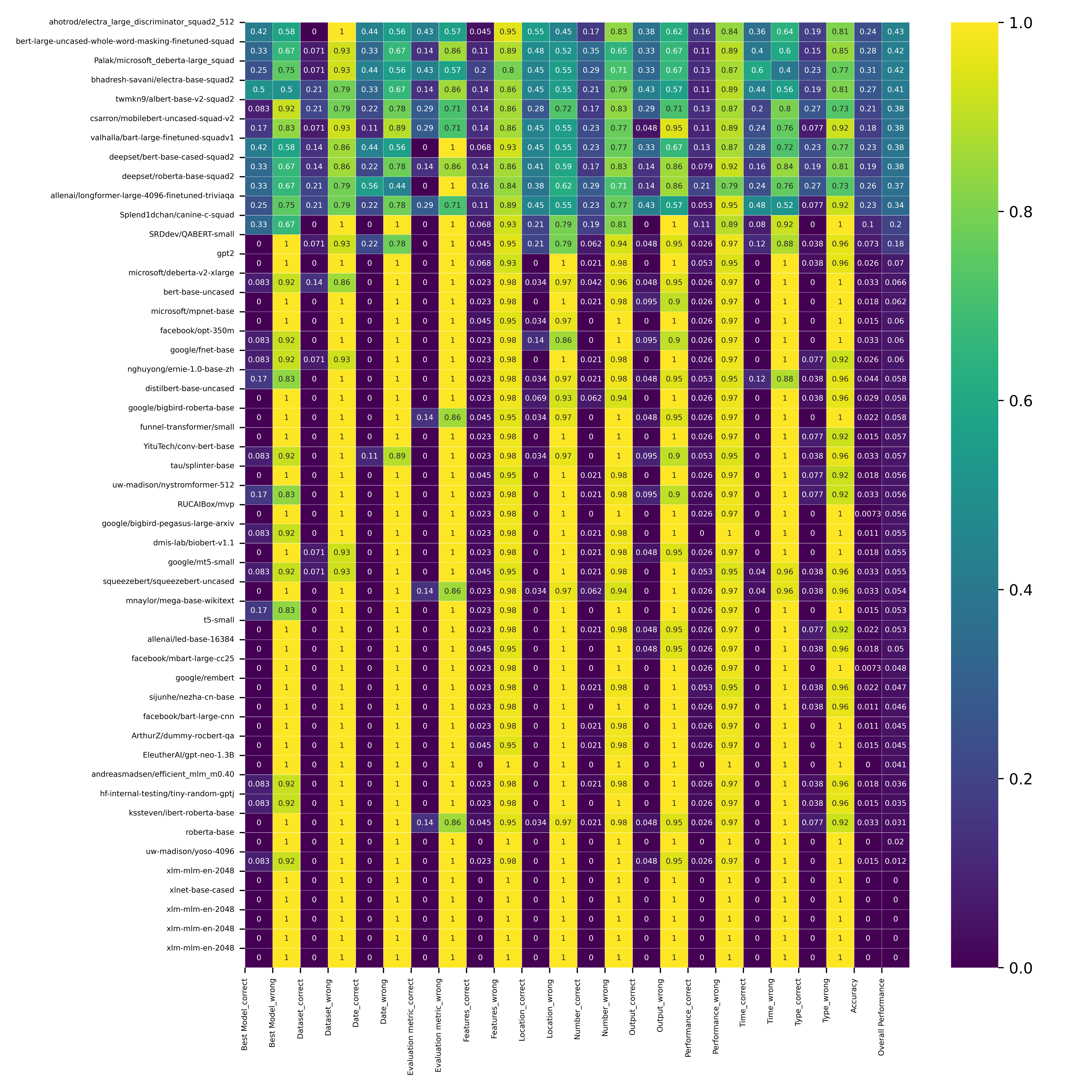}}
     \caption{The heatmap illustrates the performance of models across various question types within the JournalQA dataset. Notably, there is variation in model performance across different question types.}
\label{step11}
\end{figure}

We visualized the performance of all models through a heatmap \ref{step11}, offering insights into the efficacy of models for specific question types using research articles as plain text as context. Notably, one model can only extract some types of questions. Question types impact the model's performance, requiring various models to extract a diverse range of question types.

Table \ref{journalquestiontype} shows which model is best for which type of question time regarding journal-related question answering.

\begin{table}[!ht]
\centering
\resizebox{\columnwidth}{!}{%
\begin{tabular}{|l|l|l|}
\hline
\textbf{Index} & \textbf{Model} & \textbf{Journal question type} \\ \hline
\textbf{1} & twmkn9/albert-base-v2-squad2 & {[}'Dataset', 'Type'{]} \\ \hline
\textbf{2} & deepset/roberta-base-squad2 & {[}'Date', 'Performance'{]} \\ \hline
\textbf{3} & Palak/microsoft\_deberta-large\_squad & {[}'Evaluation metric', 'Features', 'Time'{]} \\ \hline
\textbf{4} & bhadresh-savani/electra-base-squad2 & {[}'Best Model', 'Output'{]} \\ \hline
\textbf{5} & bert-large-uncased-whole-word-masking-finetuned-squad & {[}'Number'{]} \\ \hline
\textbf{6} & ahotrod/electra\_large\_discriminator\_squad2\_512 & {[}'Location'{]} \\ \hline
\end{tabular}%
}
\caption{Table showing the best model for specific question types for the JournalQA dataset.}
\label{journalquestiontype}
\end{table}

We analyzed models that exhibited poor performance and observed that such models were designed for masked language, multi-language (cross-language) tasks, Chinese language, or other specific language. Additionally, for a significant portion of underperforming models, the corresponding documentation lacked information about the datasets used. This limitation constrained our analysis, as we could only glean insights from the limited performance of these models.

The best-performing model for each dataset and the best accuracy are shown in table \ref{finalresults}.

\begin{table}[!ht]
\centering
\resizebox{1.1\textwidth}{!}{%
\begin{tabular}{|l|l|l|l|}
\hline
\textbf{Index} & \textbf{Dataset name} & \textbf{Best Model} & \textbf{Accuracy} \\ \hline
\textbf{1} & atlas-math-sets & bhadresh-savani/electra-base-squad2 & 3.72\% \\ \hline
\textbf{2} & bioasq10b-factoid & ahotrod/electra\_large\_discriminator\_squad2\_512 & 65.92\% \\ \hline
\textbf{3} & biomedical\_cpgQA & ahotrod/electra\_large\_discriminator\_squad2\_512 & 96.45\% \\ \hline
\textbf{4} & IELTS & bert-large-uncased-whole-word-masking-finetuned-squad & 82\% \\ \hline
\textbf{5} & JournalQA & Palak/microsoft\_deberta-large\_squad & 31\% \\ \hline
\textbf{6} & QuAC & ahotrod/electra\_large\_discriminator\_squad2\_512 & 11.13\% \\ \hline
\textbf{7} & ScienceQA & twmkn9/albert-base-v2-squad2 & 24.6\% \\ \hline
\textbf{8} & Question Answer Dataset & ahotrod/electra\_large\_discriminator\_squad2\_512 & 41.6\% \\ \hline
\end{tabular}%
}
\caption{This table shows the dataset, the best-performing model for each dataset, and the final accuracy achieved by the model.}
\label{finalresults}
\end{table}

\section*{Discussion}
This comprehensive study utilized 47 context-based question-answering models across eight diverse datasets. We conducted a thorough analysis to explain the impact of dataset diversity, model size, context length, context complexity, answer length, and question type on model performance. The insights from this research provide valuable considerations for optimizing question-answering tasks.

The performance analysis identified the top 11 models as statistically significant compared to others. This finding suggests that any of these models, particularly those with less computation time, can be effectively deployed for question-answer prediction tasks. The ahotrod/electra\_large\_discriminator\_squad2\_512 model appeared as the best-performing algorithm, making it a recommended option, especially when dealing with custom datasets that differ significantly from those considered in this study. Furthermore, models trained on SQuAD v2 or SQuAD v1 datasets showed the potential for outperforming models trained on other datasets despite having similar architectures.
Examining the impact of answer length on performance revealed a decrease in model performance as the answer length increased. For tasks (medical treatment and disease name extraction) requiring accurate results, selecting a model suitable for handling long answers, such as ahotrod/electra\_large\_discriminator\_squad2\_512, is crucial.
While larger models may lead to increased computation time, for tasks where a slight decrease in accuracy is acceptable for faster results, opting for a model with a small size is advisable. We observed no specific variation in performance concerning context complexity.
Contrary to expectations, combining models based on their scores yielded negligible improvement. We recommend simultaneous training and a combination of multiple architectures during the pretraining stage to enhance model performance. This strategy allows for the concurrent updating of weights in the combined model, thereby improving its learning capability.

Furthermore, our analysis of question types within the JournalQA dataset revealed that the best-performing model varied for each question type. Therefore, a tailored approach based on question type classification is recommended. We suggest incorporating a classification head during pretraining to categorize question types, enabling selecting the best-performing model based on the specific question type.
The referenced models, datasets, and their corresponding performance implications contribute to the broader understanding of effective model selection for diverse applications, particularly in extracting answers from journal texts. The code used for the analysis conducted in this study is accessible on GitHub at the following link: \url{https://github.com/MuhammadMuneeb007/Comparative-Analysis-of-47-Context-Based-Question-Answer-Models-Across-8-Diverse-Datasets}. Additionally, the pre-and post-processed datasets can be downloaded from The Open Science Framework (OSF) using the following link: \url{https://osf.io/rd3x2/}.

\section*{Materials and Methods}
We considered eight different datasets from various sources to assess the performance of 47 question-answering models \ref{datainformation}. \textbf{Atlas-math-sets} centers around mathematical concepts related explicitly to sets and mathematical operations (addition, subtraction, multiplication, and division). \textbf{Bioasq10b-factoid} contains questions from the BioASQ challenge, which contains biomedical question answers. \textbf{Bbiomedical\_cpgQA} is tailored for question-answering tasks within the biomedical domain, explicitly focusing on Clinical Practice Guidelines (CPG). It encompasses a set of questions from medical practices and guidelines, providing a comprehensive resource for training and evaluating models in the healthcare domain. \textbf{QuAC} (Question Answering in Context) is a dataset for advancing question-answering research. Questions are posed within a conversational context, demanding the model to consider the complexities of previous dialogue. \textbf{ScienceQA} focuses on questions and answers related to scientific topics. The \textbf{Question Answer Dataset} consists of articles from Wikipedia containing context, questions, and answers. The \textbf{IELTS} (International English Language Testing System) dataset is designed to assess English language proficiency across listening, reading, writing, and speaking skills. The IELTS dataset is centered on the English language, encompassing reading comprehension with corresponding answers. We compiled the dataset of 50 questions from various online IELTS sample tests. The actual URL to the test, the passage or context, the question, and the answer to that question are available on GitHub. We considered only those IELTS questions that ended with a question mark sign. The \textbf{JournalQA} dataset is crafted from scholarly articles. Research articles contain information like the model's performance, the best-performing model, population, specific methodology, and other information that can be extracted using question-answer models. To generate such questions, we used two systematic literature reviews. The first review focused on articles related to solar power prediction \cite{Baaran2020}, while the second centered on using polygenic risk scores for predicting type 2 diabetes \cite{PadillaMartnez2020}.

\begin{table}[!ht]
\centering
\resizebox{1.1\textwidth}{!}{%
\begin{tabular}{|l|l|l|l|l|l|l|l|}
\hline
 & \textbf{Dataset} & \textbf{Link} & \textbf{Preprocessing} & \textbf{Set} & \textbf{Answer Word Threshold} & \textbf{Type} & \textbf{Number of Questions} \\ \hline
1 & \textbf{atlas-math-sets} & \url{https://huggingface.co/datasets/AtlasUnified/atlas-math-sets} & Yes & Test - First 10000 & 5 & Maths & 10000 \\ \hline
2 & \textbf{bioasq10b-factoid} \cite{Tsatsaronis2015} & \url{https://huggingface.co/datasets/legacy107/bioasq10b-factoid} & Yes & Train & 5 & Biology & 983 \\ \hline
3 & \textbf{biomedical\_cpgQA} & \url{https://huggingface.co/datasets/chloecchng/biomedical\_cpgQA} & Yes & Train & 5 & Biomedical & 535 \\ \hline
4 & \textbf{IELTS} & Collected from online resources & No & - & - & IELTS & 50 \\ \hline
5 & \textbf{JournalQA} & Collected from research articles & Yes & - & - & Science & 273 \\ \hline
6 & \textbf{QuAC} \cite{https://doi.org/10.48550/arxiv.1808.07036} & \url{https://quac.ai/} & Yes & Validation & 5 & General & 618 \\ \hline
7 & \textbf{ScienceQA} \cite{lu2022learn} & \url{https://scienceqa.github.io/} & Yes & Test & 5 & Science & 1378 \\ \hline
8 & \textbf{Question-Answer Dataset} & \url{https://www.kaggle.com/datasets/rtatman/questionanswer-dataset} & Yes & S08,S09,S09 & 5 & General & 2037 \\ \hline
\end{tabular}%
}
\caption{\textbf{Overview of Question Answer Datasets}. The first column denotes the dataset name, while the second column contains the download link. The third column indicates whether preprocessing was performed on the dataset. The information in each dataset is presented in different formats, and we transformed the information to generate context, questions, and answers for each dataset. Some datasets offer three sets (Training, Testing, and Validation), and for a specific dataset, questions were selected from a particular set, as indicated in the fourth column. Due to lengthy answers in all datasets, we considered answers having a length of less than or equal to 5 words. The sixth column classifies the dataset type (general questions, mathematical queries, or biology-related questions). The seventh column specifies the number of questions after preprocessing.}
\label{datainformation}
\end{table}

Table \ref{informationlength} displays the average context length, question length, response length, and context complexity for all datasets. These metrics provide insights into the parameters influencing the performance of CBQA models. Context complexity is calculated using the Flesch-Kincaid grade complexity formula.

\begin{table}[!ht]
\centering
\resizebox{1.1\textwidth}{!}{%
\begin{tabular}{|l|l|l|l|l|l|}
\hline
{\color[HTML]{FFFFFF} ID} & \textbf{Dataset Name} & \textbf{Average Context Length} & \textbf{Average Question Length} & \textbf{Average Response Length} & \textbf{Average Context Complexity} \\ \hline
1 & \textbf{atlas-math-sets} & 3 & 8 & 1 & -3 \\ \hline
2 & \textbf{bioasq10b-factoid} & 351 & 9 & 2 & 17 \\ \hline
3 & \textbf{biomedical\_cpgQA} & 161 & 13 & 5 & 13 \\ \hline
4 & \textbf{IELTS} & 907 & 12 & 2 & 10 \\ \hline
5 & \textbf{JournalQA} & 5568 & 9 & 2 & 11 \\ \hline
6 & \textbf{QuAC} & 84 & 6 & 2 & 11 \\ \hline
7 & \textbf{ScienceQA} & 147 & 16 & 2 & 6 \\ \hline
8 & \textbf{Question-Answer Dataset} & 5227 & 9 & 2 & 11 \\ \hline
\end{tabular}%
}
\caption{\textbf{Overview of Average Context, Answer, and Question length across all datasets.}}
\label{informationlength}
\end{table}

Table \ref{table1} shows the model name, size, and the dataset used to fine-tune the models.

\begin{table}[!ht]
\centering
\resizebox{1.1\textwidth}{!}{%
\begin{tabular}{|l|l|l|l|}
\hline
\textbf{Index} & \textbf{Model Name} & \textbf{Model Size (MB)} & \textbf{Dataset} \\ \hline
\textbf{1} & twmkn9/albert-base-v2-squad2 & 12 & SQuAD v2 \\ \hline
\textbf{2} & valhalla/bart-large-finetuned-squadv1 & 388 & SQuAD v1 \\ \hline
\textbf{3} & deepset/bert-base-cased-squad2 & 104 & SQuAD v2 \\ \hline
\textbf{4} & google/bigbird-roberta-base \cite{zaheer2021big} & 122 & Books, CC-News, Stories and Wikipedia. \\ \hline
\textbf{5} & google/bigbird-pegasus-large-arxiv \cite{zaheer2021big} & 551 & Arxiv dataset \\ \hline
\textbf{6} & dmis-lab/biobert-v1.1 & 104 & NA \\ \hline
\textbf{7} & deepset/roberta-base-squad2 & 119 & SQuAD v2 \\ \hline
\textbf{8} & Splend1dchan/canine-c-squad & 126 & NA \\ \hline
\textbf{9} & YituTech/conv-bert-base & 101 & NA \\ \hline
\textbf{10} & Palak/microsoft\_deberta-large\_squad & 387 & SQuAD v1 \\ \hline
\textbf{11} & microsoft/deberta-v2-xlarge & 844 & NA \\ \hline
\textbf{12} & distilbert-base-uncased \cite{Sanh2019DistilBERTAD} & 64 & BookCorpus and English Wikipedia \\ \hline
\textbf{13} & bhadresh-savani/electra-base-squad2 & 104 & SQuAD v2 \\ \hline
\textbf{14} & nghuyong/ernie-1.0-base-zh \cite{sun2019ernie} & 96 & Chinese \\ \hline
\textbf{15} & xlm-mlm-en-2048 \cite{lample2019cross} & 637 & Masked language modeling \\ \hline
\textbf{16} & google/fnet-base \cite{DBLP:journals/corr/abs-2105-03824} & 80 & Colossal Clean Crawled Corpus (C4) \\ \hline
\textbf{17} & funnel-transformer/small \cite{dai2020funneltransformer} & 125 & BookCorpus, English Wikipedia, Clue Web, GigaWord, and Common Crawl \\ \hline
\textbf{18} & EleutherAI/gpt-neo-1.3B \cite{gpt-neo} & 1255 & Pile \\ \hline
\textbf{19} & hf-internal-testing/tiny-random-gptj & 1 & NA \\ \hline
\textbf{20} & gpt2 & 119 & WebText \\ \hline
\textbf{21} & kssteven/ibert-roberta-base \cite{kim2021bertB20} & 119 & NA \\ \hline
\textbf{22} & allenai/led-base-16384 & 155 & NA \\ \hline
\textbf{23} & allenai/longformer-large-4096-finetuned-triviaqa & 415 & NA \\ \hline
\textbf{24} & facebook/mbart-large-cc25 & 583 & Multilingual mbart model \\ \hline
\textbf{25} & mnaylor/mega-base-wikitext & 8 & wikitext-103 \\ \hline
\textbf{26} & csarron/mobilebert-uncased-squad-v2 & 24 & SQuAD v2 \\ \hline
\textbf{27} & microsoft/mpnet-base & 105 & NA \\ \hline
\textbf{28} & google/mt5-small & 165 & 101 languages \\ \hline
\textbf{29} & RUCAIBox/mvp \cite{tang2022mvp} & 388 & NA \\ \hline
\textbf{30} & sijunhe/nezha-cn-base & 98 & Chinese \\ \hline
\textbf{31} & uw-madison/nystromformer-512 & 104 & BookCorpus and English Wikipedia \\ \hline
\textbf{32} & facebook/opt-350m \cite{zhang2022opt} & 316 & BookCorpus, CC-Stories, The Pile, Pushshift.io, and CCNewsV2 \\ \hline
\textbf{33} & bert-base-uncased \cite{DBLP:journals/corr/abs-1810-04805} & 105 & BookCorpus and English Wikipedia \\ \hline
\textbf{34} & google/rembert \cite{DBLP:conf/iclr/ChungFTJR21} & 550 & Multilingual Wikipedia data over 110 languages. \\ \hline
\textbf{35} & roberta-base \cite{DBLP:journals/corr/abs-1907-11692} & 119 & BookCorpus, English Wikipedia, CC-News, OpenWebText, and Stories \\ \hline
\textbf{36} & andreasmadsen/efficient\_mlm\_m0.40 & 339 & NA \\ \hline
\textbf{37} & ArthurZ/dummy-rocbert-qa & 112 & NA \\ \hline
\textbf{38} & tau/splinter-base \cite{ram-etal-2021-shot} & 103 & BookCorpus and English Wikipedia \\ \hline
\textbf{39} & squeezebert/squeezebert-uncased \cite{2020_SqueezeBERT} & 49 & BookCorpus and English Wikipedia \\ \hline
\textbf{40} & t5-small \cite{2020t5} & 58 & Colossal Clean Crawled Corpus (C4) \\ \hline
\textbf{41} & xlm-mlm-en-2048 \cite{lample2019cross} & 637 & Masked language modeling \\ \hline
\textbf{42} & xlnet-base-cased \cite{DBLP:journals/corr/abs-1906-08237} & 112 & XLNet model pre-trained on English language \\ \hline
\textbf{43} & uw-madison/yoso-4096 & 121 & Masked language modeling \\ \hline
\textbf{44} & SRDdev/QABERT-small \cite{QA-BERT-small} & 64 & 30k samples from the Stanford Question Answering Dataset \\ \hline
\textbf{45} & bert-large-uncased-whole-word-masking-finetuned-squad   \cite{DBLP:journals/corr/abs-1810-04805} & 320 & BookCorpus and English Wikipedia \\ \hline
\textbf{46} & facebook/bart-large-cnn \cite{DBLP:journals/corr/abs-1910-13461} & 388 & CNN Daily Mail \\ \hline
\textbf{47} & ahotrod/electra\_large\_discriminator\_squad2\_512 & 319 & SQuAD v2 \\ \hline
\end{tabular}%
}
\caption{\textbf{The table shows language models from Hugging Face included in this study.} For access to the documentation of each model, utilize the following pattern: \url{https://huggingface.co/ + Model Name} (e.g., \url{https://huggingface.co/twmkn9/albert-base-v2-squad2}). The Model Name column presents the name of each model, and the Model Size column denotes the size of the respective models. Additionally, the dataset column specifies the datasets utilized to fine-tune each model.}
\label{table1}
\end{table}

\subsection*{Methodology}
\begin{figure*}[!ht]
     \includegraphics[width=11cm,height=13cm]{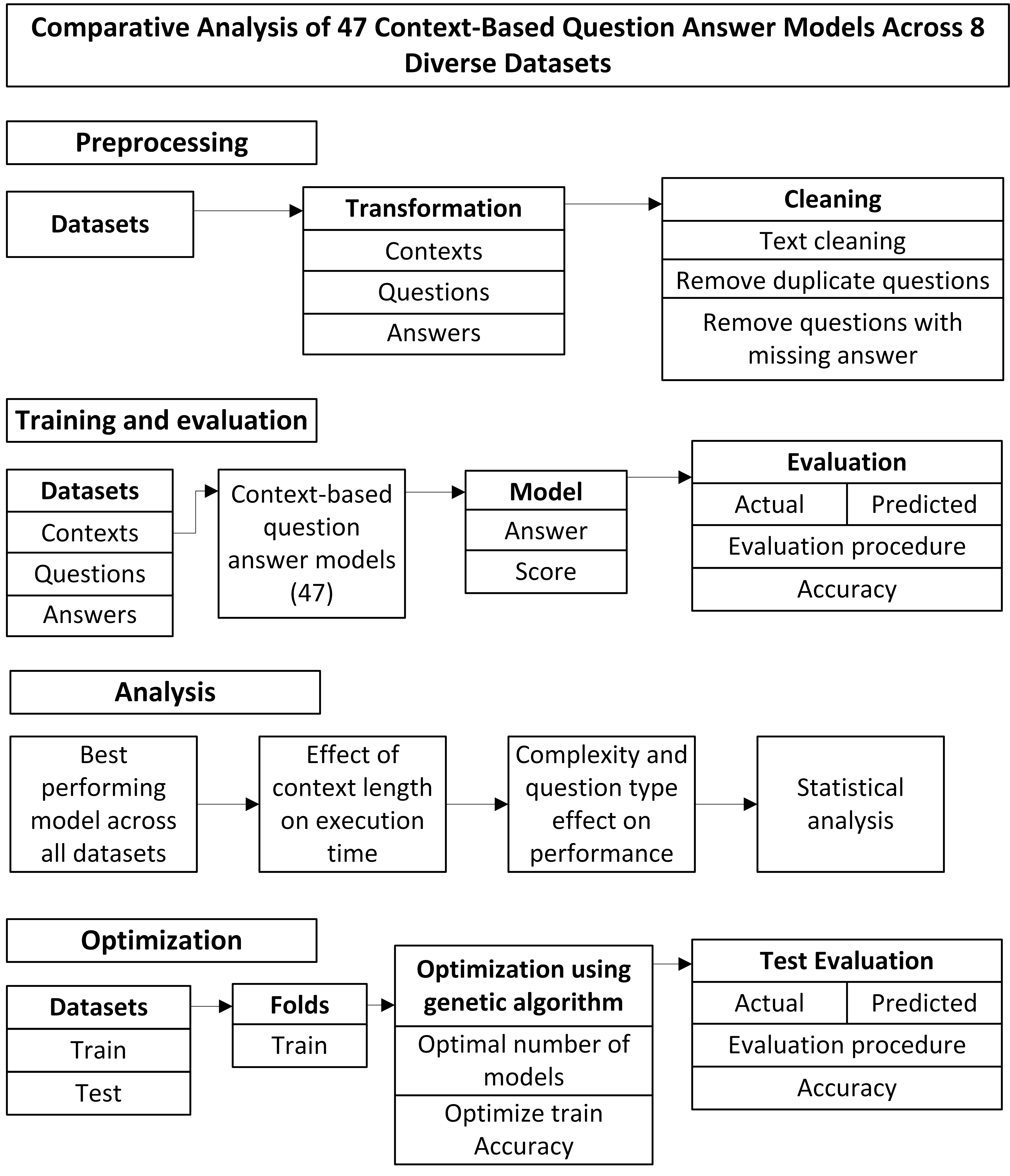}
     \caption{\textbf{The flowchart illustrates the benchmarking process for context-based question-answer models.} We processed datasets and generated a context, question, and answer for each dataset. Each dataset was passed through a context-based question-answer model, and the actual and predicted answers were compared using the evaluation procedure to calculate accuracy. We performed different kinds of analyses, including finding the best model, assessing the effect of context length on execution time, and determining the impact of question type on performance. Finally, we used a genetic algorithm to determine an optimal number and combination of models to optimize performance.}
\label{flowchart}
\end{figure*}

This section elaborates on the preprocessing and transforming datasets, the training and evaluation of models, and an ensemble-based approach to finding the optimal combination of models \ref{flowchart}.

\subsubsection*{Processing}
All downloaded datasets in various formats (Parquet or JSON) were transformed into a tabular format. Fields containing information for context-based question answering were mapped to three sets: question, context, and answer. Questions within all datasets where the answer length fell below or equal to five were retained. Text cleaning was performed on the question and context columns, replacing newline characters and stripping unnecessary whitespaces. Rows with duplicate questions, entries, or missing values were removed to ensure data accuracy and consistency.

\subsubsection*{Models Training}
Pre-trained CBQA models from Hugging Face \cite{huggingface} were utilized for question answering on processed datasets without additional fine-tuning on each dataset. The context and question columns, containing pertinent information for question answering, were extracted from all datasets and fed into the models. The model outputs, comprising both answers and their corresponding scores, were preserved for subsequent analysis. The execution time for all models was recorded due to its critical importance in performance assessments.

Table \ref{modelfrequency} shows the most occurring datasets used to train the models included in this study. SQuAD v2, BookCorpus, and English Wikipedia were among the highly used datasets to train most of the models provided by Hugging Face. A description of each model included in this study, along with the corresponding research paper, model size, and the dataset used to train each model, is shown in Table \ref{table1}.

\begin{table}[!ht]
\centering
\begin{tabular}{|l|l|l|}
\hline
 & \textbf{Dataset to Train Models} & \textbf{Frequency} \\ \hline
1 & SQuAD v2 \cite{squad} & 6 \\ \hline
2 & BookCorpus \cite{Zhu_2015_ICCV} and English Wikipedia & 6 \\ \hline
3 & Masked language modeling & 3 \\ \hline
4 & Chinese & 2 \\ \hline
5 & Colossal Clean Crawled Corpus (C4) \cite{2019t5} & 2 \\ \hline
6 & SQuAD v1 \cite{squad} & 2 \\ \hline
\end{tabular}%

\caption{This table shows the most occurring datasets used to train the models. The first column displays the dataset, and the second column shows the number of models trained using that dataset.}
\label{modelfrequency}
\end{table}

\subsubsection*{Models Evaluation}
For accuracy evaluation, we used a combination of Exact Match (EM) and text similarity to assess the performance of the CBQA models \cite{Liddy2004}. First, we performed text processing on the predicted and actual answers before evaluating the model's performance. The actual and predicted answers underwent tokenization, and we applied a series of cleaning steps, including removing leading and trailing whitespaces, translating special characters, converting text to lowercase, and eliminating English stop words. A similarity score was then calculated between the actual and predicted answers, incorporating a comprehensive set of conditions. Specifically, a similarity score of 0 was assigned when the predicted answer was null. In cases where the predicted and actual answers matched perfectly, a similarity score of 1 was assigned. We used the fuzzy-wuzzy library to compute the Levenshtein distance ratio to enhance the accuracy assessment. If the Levenshtein distance ratio exceeded 80\%, we assigned a similarity score of 1. Conversely, a similarity score of 0 was assigned when none of these conditions were met. This evaluation criterion facilitated the computation of nuanced similarity scores, offering a detailed measure of the model's accuracy.

\subsubsection*{Analysis}
After evaluating the performance of all models across various datasets, we conducted diverse analyses to understand better why specific models performed satisfactorily or inadequately. Our initial analysis focused on assessing the impact of context length and model size on the execution time of the models. We observed variations in processing speed, with some models delivering rapid results while others required a significant amount of time. We determined the statistically significant model in terms of performance and execution time across eight datasets, providing insights into the comparative effectiveness of different models. Further analyses delved into understanding the factors contributing to superior or inferior performance, including the model specifications and the datasets used for training. This exploration aimed to uncover the nuances behind model behavior.
Another crucial analysis investigated how model performance relates to the complexity and types of questions posed to it. The nature of this analysis differed for each dataset, depending on the available information in the database. To enhance overall performance, we experimented with a combination of different models using a genetic algorithm \cite{mirjalili2019genetic}.

\subsubsection*{Ensemble-Based Approach}
After obtaining individual performance metrics from each model, we tried to enhance overall performance through an ensemble approach. We aimed to leverage individual model strength and optimize overall predicting performance. Employing a 5-fold cross-validation technique, we partitioned the dataset into training (80\%) and test (20\%) sets to evaluate the effectiveness of the ensemble strategy. 
 
We evaluated 47 models, providing a diverse set for potential ensemble combinations ranging from 2 to 47. Consider a scenario where we aim to construct an ensemble comprising three models, capitalizing on the predictive capabilities of this trio. We used a genetic algorithm to select the top 3 optimal models to address this. This optimization process focused on the training set, ensuring the ensemble consisted of the best-performing models. Each model within our ensemble architecture produces an answer and an associated score. In the ensemble context, the ultimate prediction is determined by the model with the highest prediction score. The goal is to optimize the ensemble of models by selecting the combination that yields the highest accuracy. 

\clearpage

\section*{Author contributions statement}
M.M. wrote the code and generated the results. M.M. and D.A. contributed to writing and reviewing the manuscript and arranging resources for the analysis. All authors reviewed the manuscript. 

\section*{Competing interests}
The author(s) declare no competing interests.

\section*{Data availability}
 The code used for the analysis conducted in this study is accessible on GitHub at the following link: \url{https://github.com/MuhammadMuneeb007/Comparative-Analysis-of-47-Context-Based-Question-Answer-Models-Across-8-Diverse-Datasets}. Additionally, the pre-and post-processed datasets can be downloaded from The Open Science Framework (OSF) using the following link: \url{https://osf.io/rd3x2/}.

\bibliography{scientificReports}

\end{document}